\documentclass[10pt,twocolumn,letterpaper]{article}

\usepackage{cvpr}              % To produce the CAMERA-READY version
\usepackage{graphicx}
\usepackage{amsmath}
\usepackage{amssymb}
\usepackage{booktabs}

\usepackage[pagebackref,breaklinks,colorlinks]{hyperref}

\usepackage[capitalize]{cleveref}
\crefname{section}{Sec.}{Secs.}
\Crefname{section}{Section}{Sections}
\Crefname{table}{Table}{Tables}
\crefname{table}{Tab.}{Tabs.}

\def\cvprPaperID{1754} % *** Enter the CVPR Paper ID here
\def\confName{CVPR}
\def\confYear{2023}

\begin{document}

\title{How Do Deepfakes Move? \\ Motion Magnification for Deepfake Source Detection}

\author{Umur Aybars \c{C}ift\c{c}i\\
Binghamton University\\
{\tt\small uciftci@binghamton.edu}
\and
\.{I}lke Demir\\
Intel Labs\\
{\tt\small ilke.demir@intel.com}
}
\maketitle

%%%%%%%%% ABSTRACT
\begin{abstract}
   With the proliferation of deep generative models, deepfakes are improving in quality and quantity everyday. However, there are subtle authenticity signals in pristine videos, not replicated by SOTA GANs. We contrast the movement in deepfakes and authentic videos by motion magnification towards building a generalized deepfake source detector. The sub-muscular motion in faces has different interpretations per different generative models which is reflected in their generative residue. Our approach exploits the difference between real motion and the amplified GAN fingerprints, by combining deep and traditional motion magnification, to detect whether a video is fake and its source generator if so. Evaluating our approach on two multi-source datasets, we obtain 97.17\% and 94.03\% for video source detection. We compare against the prior deepfake source detector and other complex architectures. We also analyze the importance of magnification amount, phase extraction window, backbone network architecture, sample counts, and sample lengths. Finally, we report our results for different skin tones to assess the bias.
\end{abstract}

%%%%%%%%% BODY TEXT
\section{Introduction}
Since the introduction of Generative Adversarial Networks~\cite{gan} in 2014, deep generative models have been invading the domain of face generation with highly photorealistic results. With the advances in transformer and attention-based modules, the control over and the interpretability of such generators are also escalating. The recent Zelensky video~\cite{zelensky} spreading misinformation about the Russian invasion of Ukraine, or the debate about Bruce Willis' deepfake rights~\cite{bruce} are just the tip of the iceberg for a desolate digital future where we cannot trust anyone or anything we see online~\cite{ucla}. On the other hand, deepfake detection initiatives finally start to take action towards unifying the efforts~\cite{deeptrustalliance}. Although deepfake detection is less commercialized than the generation methods, there are a few industry initiatives releasing detection platforms~\cite{trustedmedia} towards a trustful future. 

Deepfake detection research has been historically investigated from two main perspectives: Blind detectors~\cite{mesonet, efficientnet, xception, inception} that try to learn the artifacts of fakery just by training on several datasets, and prior-based detectors~\cite{FakeCatcher,headpose,blink,retina,etra} where the authenticity is somehow represented by hidden signals in pristine videos. Blind detectors have the disadvantages of (1) overfitting to the datasets they are trained on and (2) being prone to adversarial attacks~\cite{sophie, Carlini2017-SSP}. Thus, our approach follows the second perspective towards more generalizable deepfake detectors, where we define the hidden watermark of being human as sub-muscular motion. Moreover, deepfake source detection is much less investigated than deepfake detection. We anticipate that these motion cues are representative enough to provide not only video authenticity, but also the generative model behind.

        \begin{figure*}[ht]
        \begin{center}
           \includegraphics[width=1\linewidth]{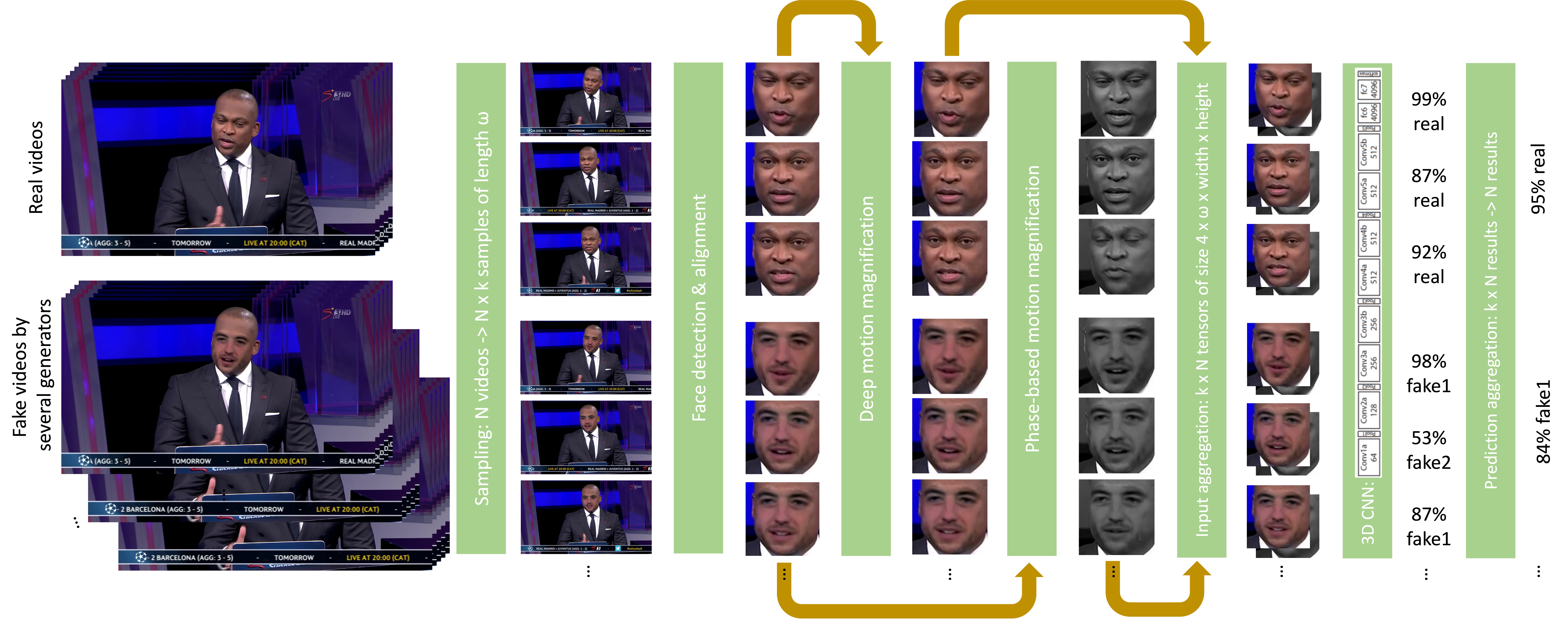}
        \end{center}
           \caption{\textbf{System Overview.} Starting with real and fake videos from several GANs, our approach selects fixed-window samples, extracts and aligns faces in those samples, applies deep and phase-based motion magnification to aligned faces, combines magnified outputs, trains a 3D CNN for source detection, and aggregates per sample predictions into video predictions to classify which GAN created each video.}
        \label{fig:overview}
        \end{figure*}
        
To reveal this motion and its projection in generative spaces of different GANs, we use motion magnification. In pristine videos, magnified motion follows the regular human motion with an emphasis, so action units and other muscles are still correlated temporally and spatially. In fake videos, we observe that the generative noise overpowers the sub-muscular motion. Thus, when the motion is magnified, the generative noise gets amplified instead of the regular human motion patterns. Our approach
\begin{itemize} 
\item analyzes the motion patterns in real and fake videos, combining traditional and deep methods; 
\item proposes a novel, robust, and generalizable deepfake source detector based on motion cues; and
\item improves both source detection and fake detection  using motion magnification, evaluated on two datasets.
\end{itemize}
        
Following the motion magnification literature, we combine traditional phase-based magnification~\cite{phasebased} which captures small temporal motions and deep magnification~\cite{deepmomag} which is more robust towards mixed motion patterns. In addition to this dual representation, we employ a 3D CNN variation to train a robust source detector which learns human motion (and its extents) in real videos and amplified generative noise in deepfakes from different source generators. Overview of our approach is depicted in Figure~\ref{fig:overview}.

We evaluate our deepfake source detector on FaceForensics++~\cite{FF++} and FakeAVCeleb~\cite{khalid2021fakeavceleb} datasets, obtaining 97.17\% and 94.03\% accuracies, respectively. In addition, we report our deepfake detection results on an in the wild dataset. We compare our source detection results against both complex blind detectors and prior-based detectors, overperforming the best one by 3.48\%. To understand the importance of motion magnification components, we conduct several experiments with different magnification levels, simple to complex backbones, different phase-windows, varying number of minimum frames, and for all skin tones.%, and on hard cases with significant illumination and motion changes. 
Finally we report the performance of the overall system and discuss how it can be deployed in current deepfake detection workflows.

\section{Related Work}%[TODO: add more recent ones]
\textbf{Deepfake Generation.}
Deepfakes have been increasing in quality and quantity since the introduction of Generative Adversarial Networks (GANs)~\cite{gan} in 2014. These approaches can (1) generate novel faces from learned distributions~\cite{choi2018stargan, stylegan, progan, mixsyn} mostly in image domain, (2) transfer or modify facial expressions, speech, identity, or mouth movements from a reference motion onto the target faces~\cite{Prajwa2020-ACMmm, 10.1145/2816795.2818056}, and (3) swap entire faces from source to target media~\cite{FaceSwap-Gan, DeepFakes, li2019faceshifter, neuraltex}. Our approach can classify videos created with any of these deepfake generation techniques and our test datasets indeed include generators from each category~\cite{FaceSwap,f2f,neuraltex,li2019faceshifter, DeepFakes, Prajwa2020-ACMmm,FaceSwap-Gan}.

\textbf{Deepfake Detection.}
As deepfakes' malevolence starts to impact the society~\cite{DeepFakeAds2, DeepFakePorn, ucla}, the arms race between generation and detection intensifies~\cite{tolosana2020deepfakesSurvey, acmsurvey}. Initial deepfake detection research focused on finding pixel-level artifacts directly from data, proposing ``blind'' detectors~\cite{mesonet,8124497,8014963,Li2020-Face-X-Ray-CVPR,shallownet,8014963,8553251,8639163,8682602,barni17,Guarnera_2020_CVPR_Workshops,Amerini2019-ICCVW}. These approaches tend to learn the specific artifacts of the datasets they are trained on, preventing their generalization and domain-transfer to any unseen video. In addition, they are more prone to be affected by adversarial attacks~\cite{sophie,Carlini2017-SSP}. 

In contrast, novel deepfake detectors aim to extract unique authenticity signals from real videos as \textit{watermarks of humans}, such as headpose~\cite{headpose}, blinking~\cite{blink}, heart-beats~\cite{FakeCatcher}, emotions~\cite{emotions}, eye and gaze properties~\cite{etra}, lighting~\cite{10.1117/12.2520546}, breathing~\cite{8553270}, and other natural, physical, or human characteristics. The consistency and correlation of these interpretable signals are broken for fake videos, so these approaches provide better generalizability as long as the GAN does not exploit the specific prior as a loss. 

\textbf{Source Detection.} The hidden artifacts of GANs, or the generative residue, have first been identified in the patterns of CNN generated images~\cite{wang2020cnngenerated}. Since then, several approaches investigate GAN fingerprints in synthetic images, with frequency analysis on 4 GANs~\cite{Yu_2019_ICCV}, in image patterns~\cite{8695364}, using latent representations~\cite{Ding2021DoesAG}, to infer model hyperparameters~\cite{asnani2021reverse-FB-MSU}, for camera attributions~\cite{Albright_2019_CVPR_Workshops}, by sensor noise~\cite{1634362}, or to poison GANs~\cite{yu2021artificial}.

Relatively less work has been proposed for videos and only one work proposes source detection on deepfakes~\cite{ijcb}. The authors classify deepfakes by their source GAN, projecting their generative residue into a biological signal domain. Our approach tackles the same problem of deepfake source detection, however we propose that motion artifacts are more representative (for pristine videos) and more fragile (for fake videos) in the context of GAN fingerprints.

\textbf{Deepfake Datasets.}
Several video datasets have been proposed for deepfake detection research, we categorize these as single-, multi-, and unknown-source datasets. Image datasets are skipped as there is no motion in single images. Single-source deepfake datasets are created by easy-access GANs and include UADFV~\cite{headpose}, DeepfakeTIMIT~\cite{vidTimid}, FaceForensics~\cite{ff}, Celeb-DF~\cite{Celeb_DF_cvpr20}, and DeeperForensics~\cite{dfor}. These datasets are crucial for deepfake detection, but not for source detection. Multi-source datasets are FaceForensics++~\cite{FF++} with 5 generators and 6K videos, DFDC~\cite{dolhansky2019deepfake} with several generators and over 100K videos, and FakeAVCeleb~\cite{khalid2021fakeavceleb} with 3 generators and 20K videos. Considering the diversity, consistency, and labeling of the datasets; we select FaceForensics++ and FakeAVCeleb datasets for training, testing, and evaluation of our approach. Finally, unknown-source deepfake datasets (i.e., in-the-wild deepfakes) have also been proposed~\cite{FakeCatcher, pu2021deepfake}, which are important for evaluating and understanding model capabilities in an in-the-wild setting. 
        
\section{Understanding Motion in Deepfakes}
Following the intuition of finding authentic and representative signals in real videos, we follow the discussion of~\cite{FakeCatcher} about biological signals. Photoplethysmography (PPG) and Ballistocardiography (BCD) signals are proposed for understanding heart beats of deepfakes, discussing that BCD extraction would require still faces, else the motion of veins would be overpowered by the actual movement. Inspired by this claim, we would like to understand the motion consistency in deepfakes.

Motion magnification is a mature research area with numerous application-specific solutions~\cite{Nowara_2021_CVPR,8777194,https://doi.org/10.1111/str.12336,app9204405}, recently extending to deep-learning-based counterparts~\cite{deepmomag}. Motion magnification has also been explored for deepfake detection recently, obtaining negative results with Euler video magnification~\cite{motion3}, without explicit motion magnification~\cite{motion1}, and using a two stage CNN+LSTM approach~\cite{motion2}. We claim that, motion discrepancy is useful not only for deepfake detection, but also for source detection, which is a different and harder problem, as the next step in the battle against deepfakes.

\begin{figure}[ht]
        \begin{center}
           \includegraphics[width=1\linewidth]{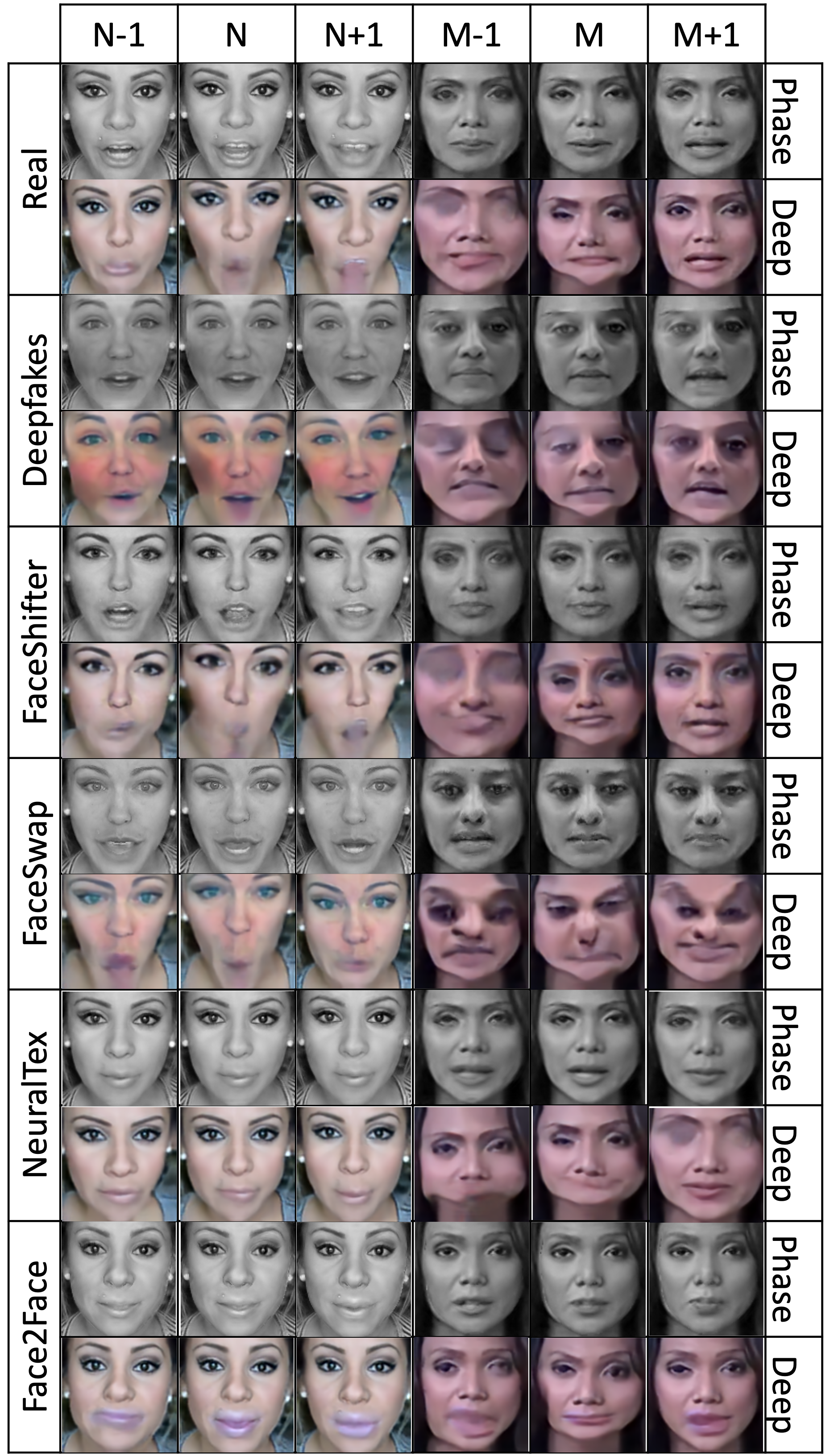}
        \end{center}
           \caption{\textbf{Motion in Deepfakes.} Each row contains 3 consecutive frames from the same 2 videos, where the motion is magnified by traditional (even rows) or deep (odd rows) methods. Real magnified frames (top 2) are followed by magnified deepfake frames from 5 different generators.}
        \label{fig:motion}
        \end{figure}

To analyze the motion in deepfakes, we first apply traditional and deep motion magnification to real and fake pairs of videos. As demonstrated in Figure~\ref{fig:motion}, the magnified motion, which is reflected as blurs in deep-motion-magnified frames, are more structural and local in real videos, whereas fake videos experience uniform blur. For the phase-based magnification, we note that the motion is reflected as an accummulation, rather than a blur. This visual observation can also be backed up by comparing the PSNR of each real and fake motion-magnified frame. Moreover, different generators (named in the first column) experience this motion dissimilarly as seen in rows 4-13, which supports our main hypothesis of \textit{``motion magnification on deepfakes reveal their source generative model, because the generative noise is amplified as opposed to real motion.''}.

\section{Motion-based Source Detection}
As depicted in Figure~\ref{fig:overview}, our approach consists of frame sampling, face processing, motion magnification, neural network training, and prediction aggregation.
\subsection{Frame Selection}
To amplify and understand the motion of the generative residue in deepfake videos, we select $k$ sample intervals of $\omega$ frames from each video for training. These samples are selected uniformly from every $(100/k)^{th}$ percentile of the video. The intuition behind this sampling is that videos in these datasets have varying lengths and we do not want any video to dominate the training process. After these fixed-sized samples are gathered, we run face detection on every frame and align detected faces to extract consistent signals. Each aligned face is fit to a 112x112 image to unify the representation. %Also, the existence of large actor motions in fake videos may overpower the motion of the residuals, so we tend to sample from still intervals.

\subsection{Motion Magnification}
As discussed in~\cite{deepmomag}, phase-based motion magnification may still perform better than deep motion magnification where temporal filters are needed to extract small motions. Thus, we combine both traditional and deep motion magnification by applying them to aligned faces of each $k$ samples of $\omega$ frames, obtaining $k\times(\omega-(t-1))\times1$ size phase-based magnification output and $k\times\omega\times 3$ size deep motion magnification output, per video. Phase-based magnification uses a sliding window of $t$ frames, thus the output is reduced in length. We merge these outputs into a tensor of $w\times h\times(\omega-(t-1))\times4$ for corresponding frames per sample, per video, as the input to our network. We left the discussion on the choices for $\omega$ and $t$ to our ablation studies in Sec.~\ref{sec:anal}.

\subsection{Network Architecture}
\label{sec:network}
Source detection task is formulated as a multi-class classification problem where $n$ fake generators in the dataset plus the originals constitute the class categories. Considering the spatio-temporal nature of our data, we attempt to use transformer-like architectures for source detection. We observe that our motion-enriched representation is powerful enough that transformers easily overfit to our data. Thus, we architect a simpler 3D convolutional neural network, similar to c3d~\cite{c3d}. Our 4D tensors are first input to 64 convolutional kernels of size 3x3x3, followed by batch norm, relu, and maxpool layers; then same block is repeated 4 times with 128, 256, 512, and 512 kernels; followed by two fully connected layers of size 4096 with 0.5 dropout. Supp. A. depicts our network. The selection of this architecture is also backed up by our experiments in Sec.~\ref{sec:anal}.

\subsection{Prediction Aggregation}
After we obtain results per sample of each video, we combine $k$ class predictions with their confidences into a video prediction. We use majority voting for this aggregation since our sample prediction accuracies are on the high side as long as there is no large motion or illumination change. Majority voting eliminates those outlier samples grounds the aggregation with respect to the possible artifacts in our videos.

\section{Results}
Our approach is implemented in Python utilizing OpenCV~\cite{opencv_library} for image processing, pytorch~\cite{pytorch} for deep learning, OpenFace~\cite{openface} for face detection and alignment, and vit-pytorch~\cite{vit-pytorch} and Efficient-3DCNN~\cite{efficient3dcnn} libraries for flexible neural network implementations. Most of the training and testing is performed on a desktop with an NVIDIA GeForce RTX 3070, where 100 epochs take a few hours to train. Applying motion magnification is the most computationally expensive part of the system, however, it is an offline task done once for training per dataset (and for each ablation study with varying parameters). Unless otherwise noted, we set $\omega=16$, $k=4$, $t=5$, and $m=2x$. Phase-based motion magnification frequency coefficients are used as-is from the original paper~\cite{phasebased} with BP 600 fps, LP 72 fps, and HP 92 fps filters. FaceForensics++~\cite{FF++} (FF) is set as the main dataset with the same 70/30 split for all evaluations \--- 700 real 700 fake videos from each 5 source GANs for training, and 300 real 300 fake videos from each 5 source GANs for testing. 

      \begin{figure}[ht]
        \begin{center}
           \includegraphics[width=1\linewidth]{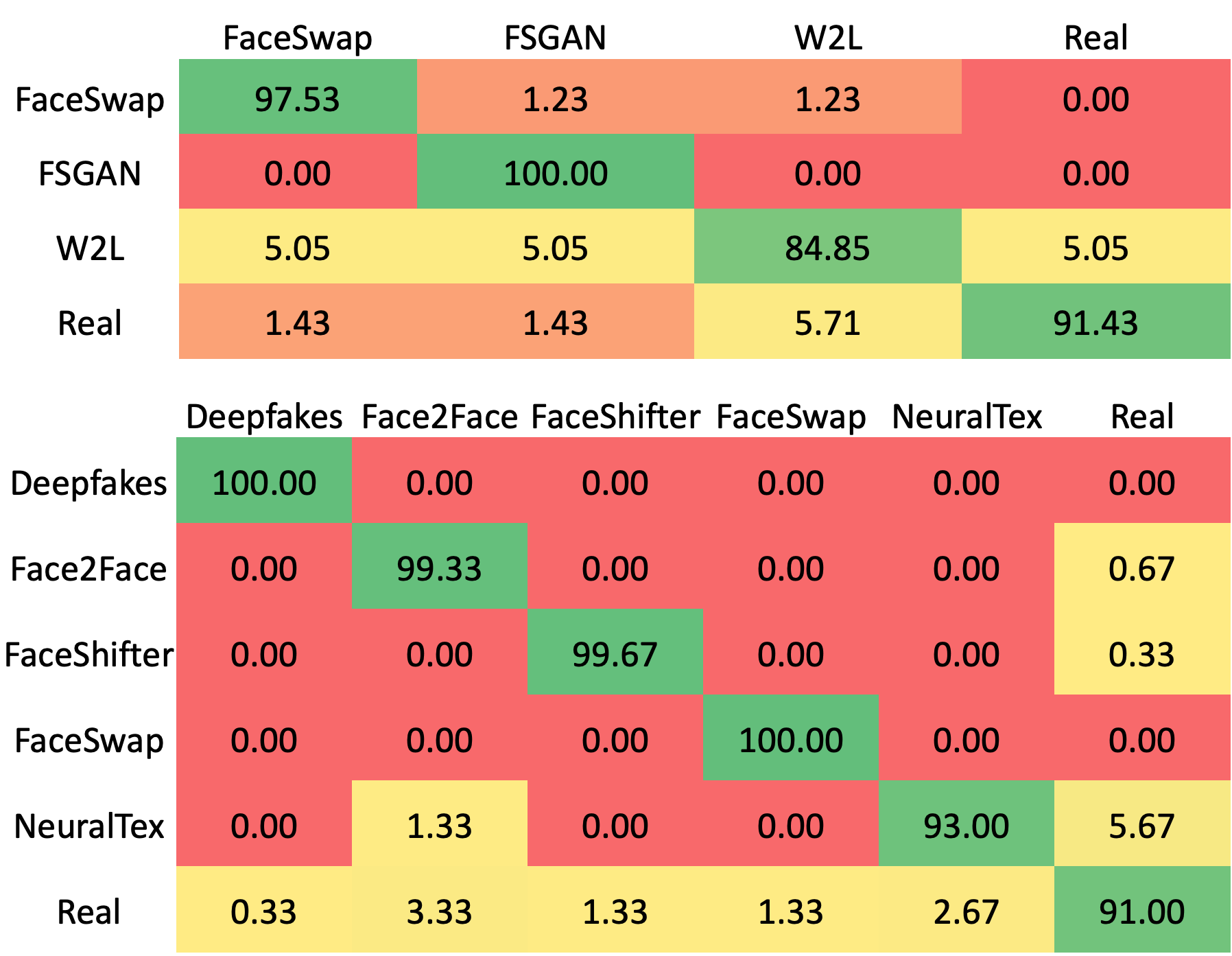}
        \end{center}
           \caption{\textbf{Source Detection Results.} Our approach obtains 94.03\% and 97.17\% overall video source detection accuracy on FAVC (top) and FF++ (bottom) datasets, respectively. }
        \label{fig:conf}
        \end{figure}
        
\subsection{Evaluation \& Comparison}
The confusion matrices in Figure~\ref{fig:conf} demonstrate our source detection accuracy per class. On FF++ dataset, we obtain 97.17\% video source detection accuracy, 95.92\% sample source detection accuracy, and 91\% fake detection accuracy. On FAVC, we obtain 94.03\% video source detection accuracy, 89.67\% sample source detection accuracy, and 91.43\% fake detection accuracy. We emphasize that, our per-class accuracies are much higher for fake classes than the real class, because the model learns the amplified motion of generative residue. In that sense, real class becomes the ``chaotic'' class where the unknown (or less confident) predictions are also pushed into.

In addition to the only other deepfake source detector in the literature~\cite{ijcb}, we compare our approach against complex network architectures used for deepfake detection, in order to emphasize the strength of our dual motion magnification representation on FF in Table~\ref{tab:Comparison1}. Our approach beats the best source detector by 3.48\% and is much simpler than the deeper networks listed, thus, it has significantly less inference time and it is more generalizable, not over-fitting to specific generators or datasets. 

%and on FakeAVCeleb~\ref{tab:Comparison2} (FAVC) in Supp. B. %Moreover, we assess the robustness of our approach by a cross-model evaluation in Table~\ref{tab:cross}, training on FF and testing on FAVC, and vice versa. 

\begin{table}[h]
\centering    
\begin{tabular}{c|c}
Models      &Source Det. Acc. 
\\\hline
ResNet50        &63.25\% \\
ResNet152      &68.92\% \\
VGG19          &76.67\% \\
Inception      &79.37\% \\ 
DenseNet201     &81.65\% \\ 
Xception        &83.50\% \\ 
PPG-based\cite{ijcb}     &93.69\% \\
%ConvLSTM & 
\hline
Ours            &\textbf{97.17\%} \\
\end{tabular}
\caption{\textbf{Comparison on FF.} Source detection accuracies of several models on FF dataset.}
\label{tab:Comparison1}
\end{table}

% \begin{table}[h]
% \centering    
% \begin{tabular}{c|c|c|c}
% Training & Testing      &Source Det. Acc.  &Fake Det. Acc.
% \\\hline
% FF & FF & & \\
% FAVC & FAVC & & \\
% FF & FAVC & & \\
% FAVC & FF & & \\
% \end{tabular}
% \caption{\textbf{Cross Model Evaluation.} Accuracies when our approach is trained and tested on FF~\cite{FF++} and FAVC~\cite{khalid2021fakeavceleb} datasets.}
% \label{tab:cross}
% \end{table}

%ijcb and other networks\\
%fake and source detection on ff++ and fakeavceleb\\
%cross model detection\\

\subsection{Analysis \& Experiments}
\label{sec:anal}
As mentioned in Section~\ref{sec:network}, we experiment with different network architectures in accordance with the characteristics of our data~\ref{tab:arch} and report both training and testing accuracies for source detection. As the motion magnified tensor representation already fortifies the generative artifacts, deeper and more complex networks tend to overfit. In order to observe this phenomenon better, we report the per-sample source detection accuracies before the aggregation step.
\begin{table}[h]
\centering    
\begin{tabular}{c|c|c}
Backbone      &Training Acc.  & Testing Acc.
\\\hline
Simple3DViT &93.11\% &53.56\%\\
3DViT       &98.60\% &45.97\%\\
CNN-LSTM    &95.76\% &44.21\%\\
ShuffleNet  &98.85\% &48.16\%\\
SqueezeNet  &99.19\% &62.65\%\\
\hline
Ours (C3D)         &99.66\% &\textbf{95.92\%}\\
\end{tabular}
\caption{\textbf{Architecture Analysis.} Training and testing accuracies with several architectures are reported to support the strength of our representation.}
\label{tab:arch}
\end{table}

In motion magnification literature, the amount of magnification is a significant parameter fine-tuned per application. Over-magnification may lead to complete loss of generative signals, as suspected to be the case in~\cite{motion3}. To investigate this claim, we experiment with several magnification coefficients for deep motion magnification and several window sizes for phase-based motion magnification in Table~\ref{tab:param}. Note that these experiments are done without the dual representation to understand the contribution of each parameter individually. The motion vectors created by the generative noise are small, thus we conclude that 2x deep magnification and 5 frame windows of phase-based magnification are the sweet spot for emphasizing the motion we pursue. As observed from these experiments, only traditional or only deep magnification is not enough to capture the generative artifacts, which underlines the contribution of our dual motion representation. 

\begin{table}[h]
\centering    
\begin{tabular}{c|c|c}
Magnification & Parameter      &Source Det. Acc.
\\\hline
%Deep &$m=1x$ (no mag.)  &90.37\%        \\
Deep &$m=2x$   &\textbf{91.54\%}  \\
Deep &$m=3x$     &86.86\%  \\
Deep &$m=4x$     &83.16\%  \\
Deep &$m=10x$     &74.90\%  \\
Phase &$t=3$      &79.88\%  \\
Phase &$t=5$     &\textbf{85.61\%}  \\
Phase &$t=7$      &81.26\%  \\
Phase &$t=10$     &82.92\%  \\
Phase &$t=16$    &64.85\%  \\\hline
Both & $m=2x$ and $t=5$ & \textbf{95.92\%}

\end{tabular}
\caption{\textbf{Motion Magnification Parameters.} We experiment with different motion magnification settings for traditional and deep components, with varying magnification coefficient ($m$) and phase-extraction interval ($t$).}
\label{tab:param}
\end{table}
         \begin{figure}[ht]
        \begin{center}
           \includegraphics[width=1\linewidth]{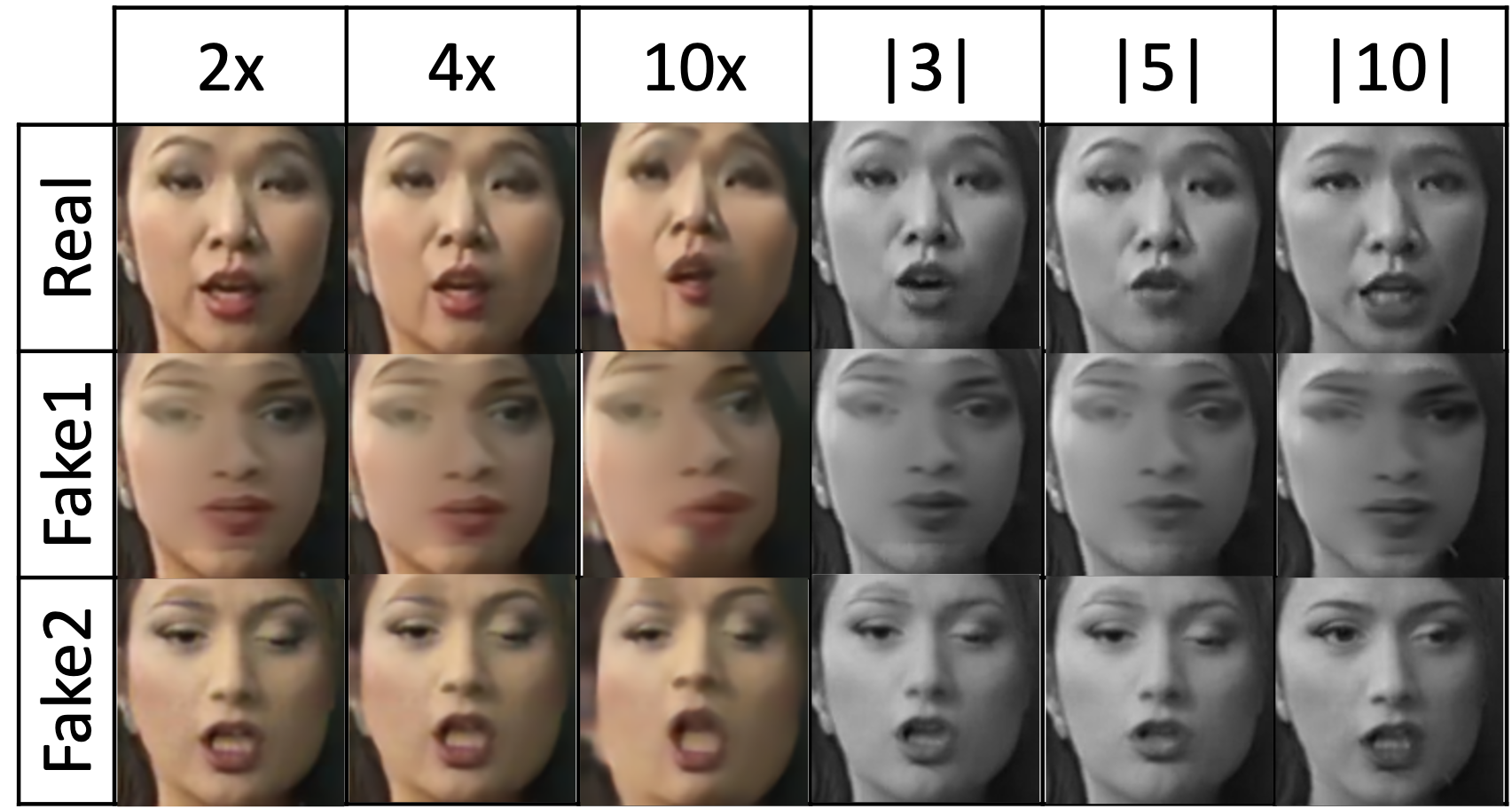}
        \end{center}
           \caption{\textbf{Magnification Parameters.} Following the experiments on different magnification parameters, we depict the effects of deep motion magnification amount $m$ (left three columns) and phase-based magnification interval $t$ (right three columns). }
        \label{fig:params}
        \end{figure}
        
In addition to this quantitative analysis, we demonstrate the effects of different parameter values in Figure~\ref{fig:params}, for a real video and two deepfakes created from it. We can observe that even for the real video, 10x magnification deteriorates the content. On the other hand, 10-frame phase extraction tends to converge to a mean image of the video, which is not useful either for capturing small motions. Based on these observations and the experiments in Table~\ref{tab:param}, we conclude with $m=2x$ and $t=5$ values.

As the last experiment, we want to detect and mitigate any possible racial or gender bias in our dataset or in our algorithm. To that end, we use the labels in FAVC dataset to report per gender and per skin tone source detection accuracies. We observe that the largest discrepancy in accuracies is between Asian women and American men, with 84.21\% and 97.44\% accuracies. We suspect that this difference may rise from the fact that deepfake generators are not creating such faces with the same fidelity, thus, the detection results are also skewed. We leave further analysis as future work.

\begin{table}[h]
\centering    
\begin{tabular}{c|c|c|c}
Skin Tone& Gender        & Sample Acc.      &Video Acc.
\\\hline
African American&Men     &79.58\%&89.19\% \\
African American&Women   &96.56\%&93.59\% \\
American&Men            &95.63\%&97.44\% \\
American&Women          &89.79\%&94.74\% \\
Asian&Men               &85.76\%&89.74\% \\
Asian&Women             &84.25\%&84.21\% \\
European&Men            &86.46\%&92.11\% \\
European&Women          &93.12\%&94.87\% \\
Indian&Men              &90.83\%&94.87\% \\
Indian&Women            &81.94\%&87.18\% \\

\end{tabular}
\caption{\textbf{Gender and Skin Tone Analysis.} Per sample and per video source detection accuracies for 5 skin tones and 2 genders.}
\label{tab:skintone}
\end{table}

We also experiment with varying number of samples per video ($k$) and changing number of frames in a sample ($\omega$). Considering the accuracy, speed, and memory requirements, we end up selecting $k=4$ and $\omega=16$. In Table~\ref{tab:K} we document experiments with $k=\{1,2,3,4\}$, concluding that $k=4$ is more informative and creates a more diverse dataset, increasing the accuracy. Larger values have were very incremental contributions, so $k=4$ has the optimum performance.
\begin{table}[h]
\centering    
\begin{tabular}{c|c}
$k$ Value  &FFPP Video Acc.
\\\hline
1&    95.72\% \\
2&   94.83\% \\
3&    96.88\% \\
4&    \textbf{97.17\%} \\
\end{tabular}
\caption{\textbf{Sample Size Analysis.} $k$ samples per video affects the accuracy. After $k=4$, the contribution is almost constant.}
\label{tab:K}
\end{table}

% \begin{table}[h]
% \centering    
% \begin{tabular}{c|c}
% Parameter      &Source Det. Acc.
% \\\hline
% $k=2$ &\\
% $k=4$ &\\
% $k=8$ &\\
% $\omega=8$ &\\
% $\omega=16$ &\\
% $\omega=32$ &\\

% \end{tabular}
% \caption{\textbf{Magnification Parameters.} We experiment with different motion magnification settings for traditional and deep components.}
% \label{tab:param}
% \end{table}

%architecture search\\
%magnification amount for both\\
%nb of frames and nb of samples\\
%without merging
\section{Conclusion and Future Work}
Following other questions about deepfakes, such as their emotions~\cite{emotions}, gazes~\cite{etra}, and hearts~\cite{ijcb}, we ask \textit{``How do deepfakes move?''}. We propose that motion magnification emphasizes the generative artifacts in deepfakes, which can be used for source detection. Combining deep and phase-based motion magnification, we build a motion-based source detection network, achieving accuracies higher than existing source detectors. We support our observations and choices with ablation studies and experiments.

In the battle against deepfakes, we believe that source detection plays a crucial role for continuous deployment and integration of detectors into trusted platforms. Emergence of novel generators as well as tracking the malevolent uses of current ones are enabled by source detection, to timely prevent deepfakes causing catastrophic events~\cite{zelensky}. In future, we would like to explore motion in deepfakes in a multi-modal setting, correlating sound, speech, gaze, and gesture signals with their motion.

%%%%%%%%% REFERENCES
{\small
\bibliographystyle{ieee_fullname}
\bibliography{MixSynBib}
}

\end{document}